\documentclass[10pt,technote,compsoc]{IEEEtran}

\ifCLASSOPTIONcompsoc
  \usepackage[nocompress]{cite}
\else
  \usepackage{cite}
\fi
\ifCLASSINFOpdf
\else
\fi

\usepackage{amsmath,amsfonts}
\usepackage{algorithmic}
\usepackage{array}
\usepackage[caption=false,font=normalsize,labelfont=sf,textfont=sf]{subfig}
\usepackage{textcomp}
\usepackage{stfloats}
\usepackage{url}
\usepackage{verbatim}
\usepackage{graphicx}
\usepackage{graphicx} 
\usepackage{amsmath}
\usepackage{amsfonts}
\usepackage{multirow}
\usepackage{color,xcolor}
\usepackage{bbding}
\usepackage{graphicx}
\usepackage{amsmath}
\usepackage{amssymb}
\usepackage{booktabs}

\usepackage{enumitem}
\usepackage{multirow}

\usepackage{xcolor}
\usepackage{rotating}
\usepackage{threeparttable}

\usepackage{algorithm}
\usepackage{algorithmic}

\usepackage{newfloat}
\usepackage{listings}
\DeclareCaptionStyle{ruled}{labelfont=normalfont,labelsep=colon,strut=off} 
\floatstyle{ruled}
\newfloat{listing}{tb}{lst}{}
\floatname{listing}{Listing}

\usepackage[colorlinks,
            linkcolor=red,       
            anchorcolor=blue,  
            citecolor=blue,        
            ]{hyperref}

\usepackage{amsthm} 

\usepackage{orcidlink}

\begin{document}
%
\title{Unveiling and Mitigating Generalized Biases of DNNs through the Intrinsic Dimensions of Perceptual Manifolds}
%
%
%
%

\author{Yanbiao Ma~\orcidlink{0000-0002-8472-1475},
        Licheng Jiao~\orcidlink{0000-0003-3354-9617},~\IEEEmembership{Fellow,~IEEE,}
        Fang Liu~\orcidlink{0000-0002-5669-9354},~\IEEEmembership{Senior Member,~IEEE,}
        Lingling Li~\orcidlink{0000-0002-6130-2518},~\IEEEmembership{Senior Member,~IEEE,}
        Wenping Ma~\orcidlink{0000-0001-8872-2195},~\IEEEmembership{Senior Member,~IEEE,}
        Shuyuan Yang~\orcidlink{0000-0002-4796-5737},~\IEEEmembership{Senior Member,~IEEE,}
        Xu Liu~\orcidlink{0000-0002-8780-5455},~\IEEEmembership{Member,~IEEE,}
        Puhua Chen~\orcidlink{0000-0001-5472-1426},~\IEEEmembership{Senior Member,~IEEE}

\thanks{This work was supported in part by the Key Scientific Technological Innovation Research Project of the Ministry of Education, the Joint Funds of the National Natural Science Foundation of China (U22B2054), the National Natural Science Foundation of China (62076192, 61902298, 61573267, 61906150, and 62276199), the 111 Project, the Program for Cheung Kong Scholars and Innovative Research Team in University (IRT 15R53), the Science and Technology Innovation Project from the Chinese Ministry of Education, the Key Research and Development Program in Shaanxi Province of China (2019ZDLGY03-06), and the China Postdoctoral Fund (2022T150506). 

\emph{(Corresponding author: Licheng Jiao.)}}
\thanks{The authors are with the Key Laboratory of Intelligent Perception and Image Understanding of the Ministry of Education of China, International Research Center of Intelligent Perception and Computation, School of Artificial Intelligence, Xidian University, Xian 710071, China (e-mail: ybmamail@stu.xidian.edu.cn; lchjiao@mail.xidian.edu.cn).}
}
%
%

\markboth{TPAMI - Paper Submission}%
{Shell \MakeLowercase{\textit{et al.}}: Bare Demo of IEEEtran.cls for Computer Society Journals}

\IEEEtitleabstractindextext{%
\begin{abstract}
Building fair deep neural networks (DNNs) is a crucial step towards achieving trustworthy artificial intelligence. Delving into deeper factors that affect the fairness of DNNs is paramount and serves as the foundation for mitigating model biases. However, current methods are limited in accurately predicting DNN biases, relying solely on the number of training samples and lacking more precise measurement tools. Here, we establish a geometric perspective for analyzing the fairness of DNNs, comprehensively exploring how DNNs internally shape the intrinsic geometric characteristics of datasets—the intrinsic dimensions (IDs) of perceptual manifolds, and the impact of IDs on the fairness of DNNs. Based on multiple findings, we propose Intrinsic Dimension Regularization (IDR), which enhances the fairness and performance of models by promoting the learning of concise and ID-balanced class perceptual manifolds. In various image recognition benchmark tests, IDR significantly mitigates model bias while improving its performance.
\end{abstract}

}

\maketitle

\IEEEdisplaynontitleabstractindextext

\IEEEpeerreviewmaketitle

\section{Introduction}
\label{sec:introduction}

\begin{figure}[t]
\begin{center}
\includegraphics[width=\linewidth]{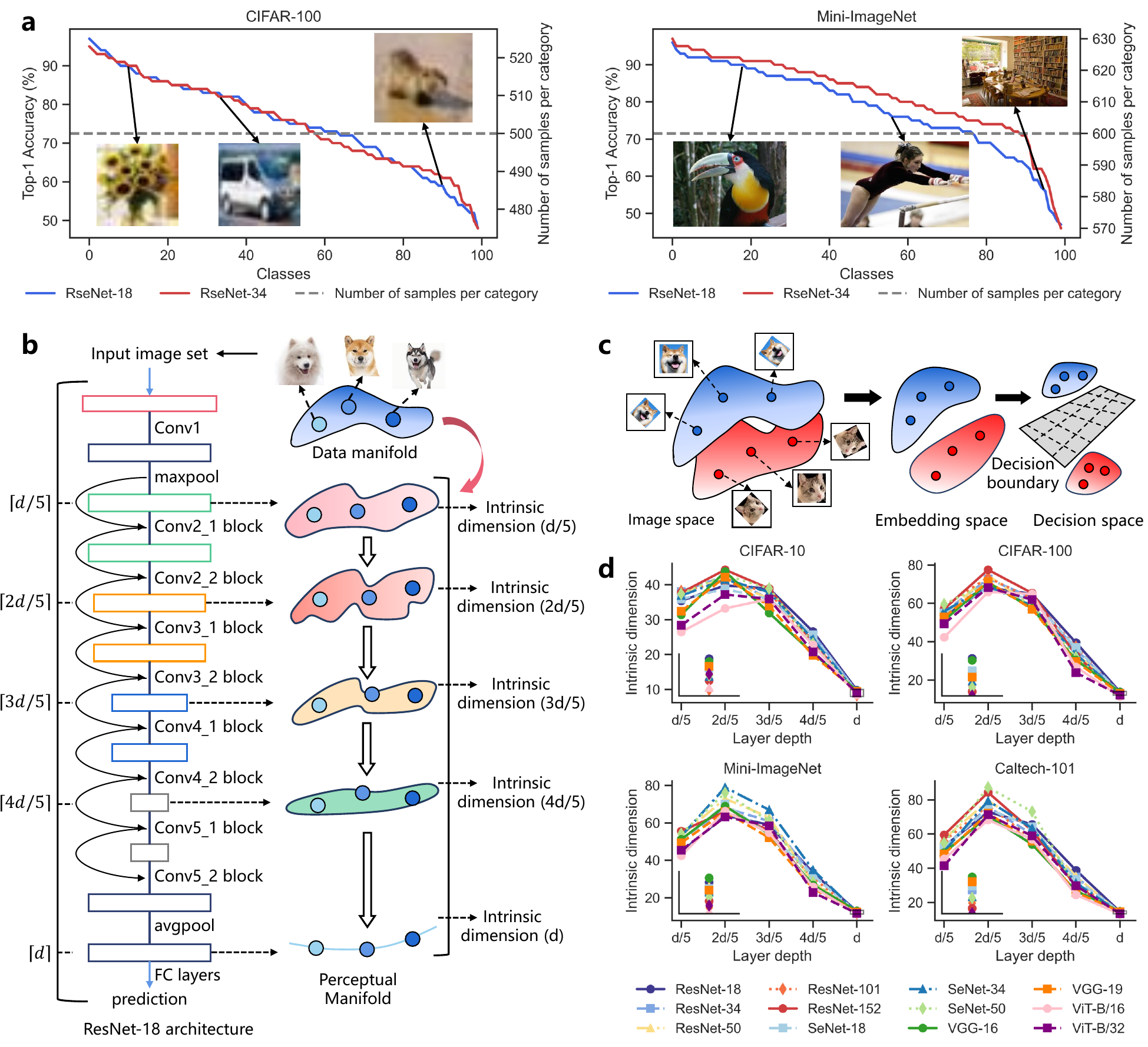}
\end{center}
\vspace{-0.15in}
\caption{\textbf{a} On a dataset containing the same number of samples per category, the trained model still has a significant bias. \textbf{b} Geometric Perspectives and Quantitative Results of Information Compression and Classification by DNN, with d representing the depth of the DNN. The process of information compression by DNN can be understood as gradually obtaining low-dimensional perceptual manifolds along the internal layers of the DNN. Low-dimensional perceptual manifolds benefit the performance of subsequent tasks. \textbf{c} Each category corresponds to a data manifold, which is mapped to perceptual manifolds in the DNN. As the depth of the DNN increases, the perceptual manifolds gradually separate from each other and the dimensions decrease, facilitating the correct identification of each category by the model. \textbf{d} Data manifolds are constructed on a dataset basis rather than by category. As the depth of the DNN increases, the intrinsic dimensions of perceptual manifolds corresponding to four benchmark image datasets show a trend of initially increasing and then decreasing.}
\label{fig1}
\vskip -0.1in
\end{figure}

\IEEEPARstart{D}{eep} neural networks (DNNs) have demonstrated unprecedented capabilities in solving complex problems and handling large-scale data sets \cite{naturehuawei,tpamiasp,tpami3}, with large models \cite{blip2,clip,tpami2} based on these networks becoming the forefront of artificial intelligence research. While striving for high performance, fairness and bias in models have emerged as key factors affecting the practical application of artificial intelligence \cite{NMI4,ACM1,tpami4}. Although previous studies have tended to attribute model bias to imbalances in sample sizes \cite{samplenumber2,samplenumber3,samplenumber4,ma1}, recent research \cite{sinha2022class,ma3,ma4} has revealed that biases still exist in models even when sample sizes are balanced (see Fig.\ref{fig1}a). This indicates that deeper factors are affecting the fairness of models. Therefore, exploring other intrinsic properties of data that affect model fairness, and developing more universal and comprehensive methods for predicting model biases, constitute an important and challenging task that is crucial for enhancing the fairness of models.

Natural images usually follow the manifold distribution law: data of one category typically cluster near a low-dimensional manifold embedded within a high-dimensional space \cite{science,Guxainfeng,ma5}. The data manifold formed by the image embeddings generated within the internal layers of DNNs is called the perceptual manifold \cite{ma4}. If the collection of images for each category is considered a data manifold, then the process of data classification can be understood as effectively untangling, reducing dimensions, and separating these perceptual manifolds corresponding to different categories through the layers of the DNN \cite{ma4,nc1} (as shown in Fig.\ref{fig1}b and \ref{fig1}c). From this geometric perspective, we hypothesize that if the data manifolds corresponding to certain categories are more complex geometrically, the DNN may encounter greater challenges in simplifying and reducing the dimensions of these perceptual manifolds, resulting in the perceptual manifolds produced by the last layers of the DNN still having relatively large intrinsic dimension (ID) \cite{ID,ID_MLE}. The differences in the intrinsic dimensions of perceptual manifolds across categories could lead to inconsistencies in the recognition capabilities of the classifiers at the end of the DNN for different categories, thereby affecting the fairness of the model.

In this study, we first comprehensively explore the impact of the ID of the perceptual manifold generated by DNN on model performance and fairness. \textbf{Specifically, our research unfolds in the following areas:} \textbf{(1)} How the ID of the dataset's corresponding perceptual manifold affects the overall performance of the model. \textbf{(2)} How the learning process of the model affects the ID of the perceptual manifold. \textbf{(3)} Whether there is a correlation between the ID differences of the perceptual manifolds corresponding to different categories and model fairness. \textbf{(4)} How the model's learning process affects the imbalance of IDs among the perceptual manifolds corresponding to categories. A significant finding is that the imbalance in ID among class perceptual manifolds is significantly related to model fairness. Based on multiple observations, we propose \textbf{Intrinsic Dimension Regularization (IDR)}, which enforces the model to learn more concise and ID-balanced perceptual manifolds by adding additional constraints to the existing optimization objective. This regularization aims to alleviate model bias and enhance overall performance. Simultaneously, we devise a training strategy to efficiently utilize IDR, addressing the challenge of estimating intrinsic dimensions when embeddings of all samples are not accessible in each iteration. Experiments on multiple benchmark image datasets, both balanced and unbalanced, demonstrate that \textbf{IDR} can significantly mitigate model biases while also enhancing the overall performance of the model. We hope this work will facilitate a deeper understanding of the underlying mechanisms of deep learning while offering practical solutions to mitigate potential biases in artificial intelligence systems.

\section{Estimation of the Intrinsic Dimension}
Given a set of embeddings $Z=[z_1,\dots,z_m]\in \mathbb{R}^{p \times m}$ corresponding to an image dataset, $Z$ is typically distributed near a low-dimensional perceptual manifold $M$ embedded in the $p$-dimensional space, akin to a two-dimensional plane in three-dimensional space. The intrinsic dimension $ID(M)$ of the perceptual manifold is such that $d<p$. A higher intrinsic dimension indicates a more complex perceptual manifold. The following describes how to use TLE to estimate the intrinsic dimension of the perceptual manifold formed by $Z=[z_1,\dots,z_m]\in \mathbb{R}^{p \times m}$.

The primary method for estimating intrinsic dimension involves analyzing the distribution of distances between each point in the dataset and its neighboring points, and then estimating the dimensionality of the local space based on the rate of growth of distances or other statistic. Assuming that the distribution of samples is uniform within a small neighborhood, and then uses a Poisson process to simulate the number of points discovered by random sampling within neighborhoods of a given radius around each sample \cite{ID_MLE}. Subsequently, by constructing a likelihood function, the rate of growth in quantity is associated with the surface area of a sphere. Given any embedding $z_i$ in the dataset and its set of $k$ nearest neighbors $V$, the Maximum Likelihood Estimator (MLE) of the local intrinsic dimensiona at $z_i$ is given by: $ID_{MLE}(z_i)=-(\frac{1}{k}\sum_{j=1}^{k}ln\frac{r_j(z_i)}{r_k(z_i)})^{-1},$ where $r_j(z_i)$ represents the distance between $z_i$ and its $j$-th nearest neighbor. TLE \cite{TLE} no longer assumes uniformity of sample distribution in local neighborhoods, thus closer to the true data distribution. It assumes the local intrinsic dimension to be continuous, thereby utilizing nearby sample points to stabilize estimates at $z_i$. Specifically, the estimate of the intrinsic dimension at $z_i$ using TLE is given by 
\begin{equation}
\begin{split}
ID_{TLE}(z_i) = -\Bigg(&\frac{1}{\left | V_* \right |^2}\sum_{\substack{v,w\in V_*\\ v\neq w}}\Bigg[\ln\frac{d_{z_i}(v,w)}{r_k(z_i)} \\
&+\ln\frac{d_{z_i}(2z_i-v,w)}{r_k(z_i)}\Bigg]\Bigg)^{-1},
\end{split}
\nonumber
\end{equation}

where $V_* = V \cup \{z_i\}$, and $d_{z_i}(v,w)$ is defined as $(r_k(z_i)(w-v) \cdot (w-v)) / (2(z_i-v) \cdot (w-v))$. Furthermore, the global intrinsic dimensionality of the perceptual manifold is estimated as the average of local intrinsic dimensionalities: $$ID_{TLE}(Z)=\frac{1}{m}\sum_{i=1}^{m}ID_{TLE}(z_i).$$

In this study, we employed two intrinsic dimension (ID) estimation methods: TLE and $ID(Z)=\frac{(tr(ZZ^T))^2}{tr((ZZ^T)^2)}$ \cite{litwin2017optimal}. While TLE was used to estimate the intrinsic dimension of perceptual manifolds in most experiments, \( ID(Z) \) was specifically adopted within the intrinsic dimension regularization. Although TLE and \( ID(Z) \) differ in their computational approach, both methods fundamentally aim to quantify the geometric complexity of data distributions in high-dimensional space. TLE is particularly effective in providing a precise estimation of local geometric structures; however, it involves non-linear and non-smooth operations, leading to relatively low computational efficiency and insufficient differentiability. These characteristics make it unsuitable for direct inclusion in the loss function during the optimization process. Conversely, \( ID(Z) \) offers a simpler and smoother design, making it more appropriate for use in intrinsic dimension regularization during optimization. The decision to employ these two distinct estimation methods allows us to balance accuracy in experimental analysis with differentiability requirements in model optimization. We will provide the code for implementing intrinsic dimension estimation at \url{https://github.com/mayanbiao1234/Geometric-metrics-for-perceptual-manifolds}.

\section{Results}

We trained classical deep neural networks \cite{vgg,resnet,vit} on multiple benchmark image datasets \cite{cifar,imagenet,caltech101} for experimental exploration. Each model underwent $10$ training sessions with different random seeds. All experimental results represent the average of models trained with different initialization parameters.

\subsection{Perceptual Manifold in Deep Neural Network}

In the neural system, when neurons receive stimuli from the same category with different physical features, a perceptual manifold is formed \cite{Perceptual_Manifold2,ma4}. The formation of perceptual manifolds helps the neural system to perceive and process objects of the same category with different features distinctly. Recent studies \cite{Perceptual_Manifold1,Perceptual_Manifold3} have shown that the response of deep neural networks to images is similar to human vision and follows the manifold distribution law. Specifically, embeddings of natural images are distributed near a low-dimensional manifold embedded in a high-dimensional space. Given a set of data $X=[x_1,\dots, x_m]$, and a trained deep neural network, $Model = \{f(x,\theta_1), g(z,\theta_2)\}$, where $f(x,\theta_1)$ and $g(z,\theta_2)$ represent the representation network and classifier of the model, respectively. The representation network extracts $p$-dimensional embeddings $Z=[z_1,\dots, z_m]\in \mathbb{R}^{p\times m}$ for $X$, where $z_i=f(x_i,\theta_1) \in \mathbb{R}^p$. The point cloud manifold formed by the set of embeddings $Z$ is referred to as the perceptual manifold in deep neural networks. In the following sections, the intrinsic dimension of the perceptual manifold are all estimated using \textbf{TLE} \cite{TLE} if not otherwise specified. We describe the TLE in the Methods section.

\begin{figure}[tb]
\begin{center}
\includegraphics[width=\linewidth]{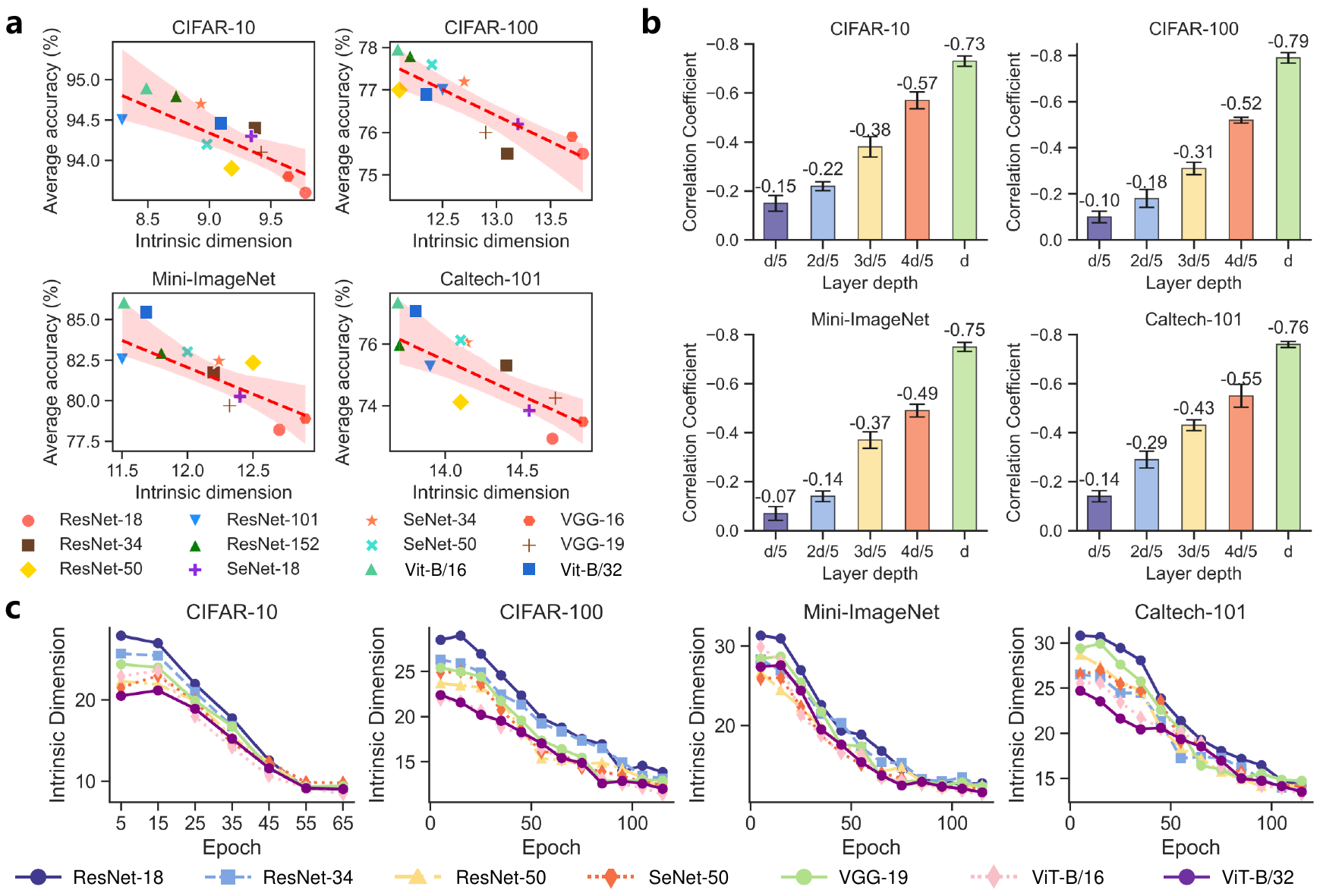}
\end{center}
\vskip -0.05in
\caption{\textbf{The Relationship between Intrinsic Dimensions of Perceptual Manifolds and Model Performance, and the Learning Process.} \textbf{a} The intrinsic dimensions of perceptual manifolds generated by the last hidden layer of DNN exhibit a negative correlation with the overall performance of the DNN across four benchmark image datasets. \textbf{b} The correlation between the intrinsic dimensions of perceptual manifolds generated by different layers of DNN and the overall performance of the DNN. Across the four benchmark datasets, there is an increasing negative correlation between the intrinsic dimensions of perceptual manifolds and the overall performance of the DNN as the depth increases. \textbf{c} As training progresses, the intrinsic dimensions of perceptual manifolds corresponding to datasets generated by multiple benchmark networks gradually decrease.}
\label{fig2}
\end{figure}

\subsection{The Intrinsic Dimension of Perceptual Manifold Exhibits Negative Correlation with DNN Performance}

DNNs compress information through hierarchical feature extraction \cite{nc1,nc2, ID_DNN1}, thereby forming lower-dimensional image representations in the last hidden layer. As shown in Fig.\ref{fig1}b, we utilized a well-trained DNN to extract image embeddings from the entire dataset at different layers and quantified the variation of the IDs of the perceptual manifolds with increasing depth of layers (Fig.\ref{fig1}d). The results reveal a pattern where the IDs of the perceptual manifolds initially increase and then decrease with the deepening of layers, rather than exhibiting a monotonic decrease. This phenomenon can be explained as DNNs, in the shallow layers, untangling the coupling between perceptual manifolds corresponding to different categories by increasing dimensions \cite{Guxainfeng}. Please note that the perceptual manifold referred to in this section is constructed from the entire dataset.

The compressive ability of DNNs may be associated with their performance \cite{ID_DNN1,ID_DNN2}. We conducted a more comprehensive and in-depth study from a geometric perspective. Initially, we evaluated the overall performance of all well-trained models on the test set. Then, we extracted embeddings of image sets from the last hidden layer of DNNs and estimated the corresponding IDs of the perceptual manifolds. Fig.\ref{fig2}a illustrates the distribution of model performance against the IDs of the perceptual manifolds generated. \textbf{We found that:} \textbf{(1)} Lower ID of the perceptual manifold corresponds to higher accuracy of model on the test set. \textbf{(2)} With the increasing complexity of network structures, the ID of the generated perceptual manifold tends to be lower. These findings suggest that DNNs capable of generating perceptual manifolds with lower intrinsic dimensions exhibit stronger feature extraction and integration capabilities. This provides a novel geometric constraint or objective for model training and optimization, namely, enhancing model performance by reducing the dimension of the perceptual manifold.

We further explored whether the IDs of perceptual manifolds from other layers of DNNs could predict the models' performance. Assuming the depth of the model is $d$, we extracted image embeddings from the $\lceil d/5 \rceil$, $\lceil 2d/5 \rceil$, $\lceil 3d/5 \rceil$, $\lceil 4d/5 \rceil$ and $\lceil d \rceil$ layers of model. Fig.\ref{fig2}b illustrates the Pearson correlation coefficients (PCCs) between the IDs of perceptual manifolds generated from different layers of the model and the model accuracy. The results indicate that with the deepening of network layers, there is an increasingly strong correlation between the IDs of perceptual manifolds and the overall performance of the model. This phenomenon provides more specific guidance on the potential application of intrinsic dimension as a geometric constraint, particularly applied to the last hidden layer.

\begin{figure*}[tb]
\begin{center}
\includegraphics[width=0.93\linewidth]{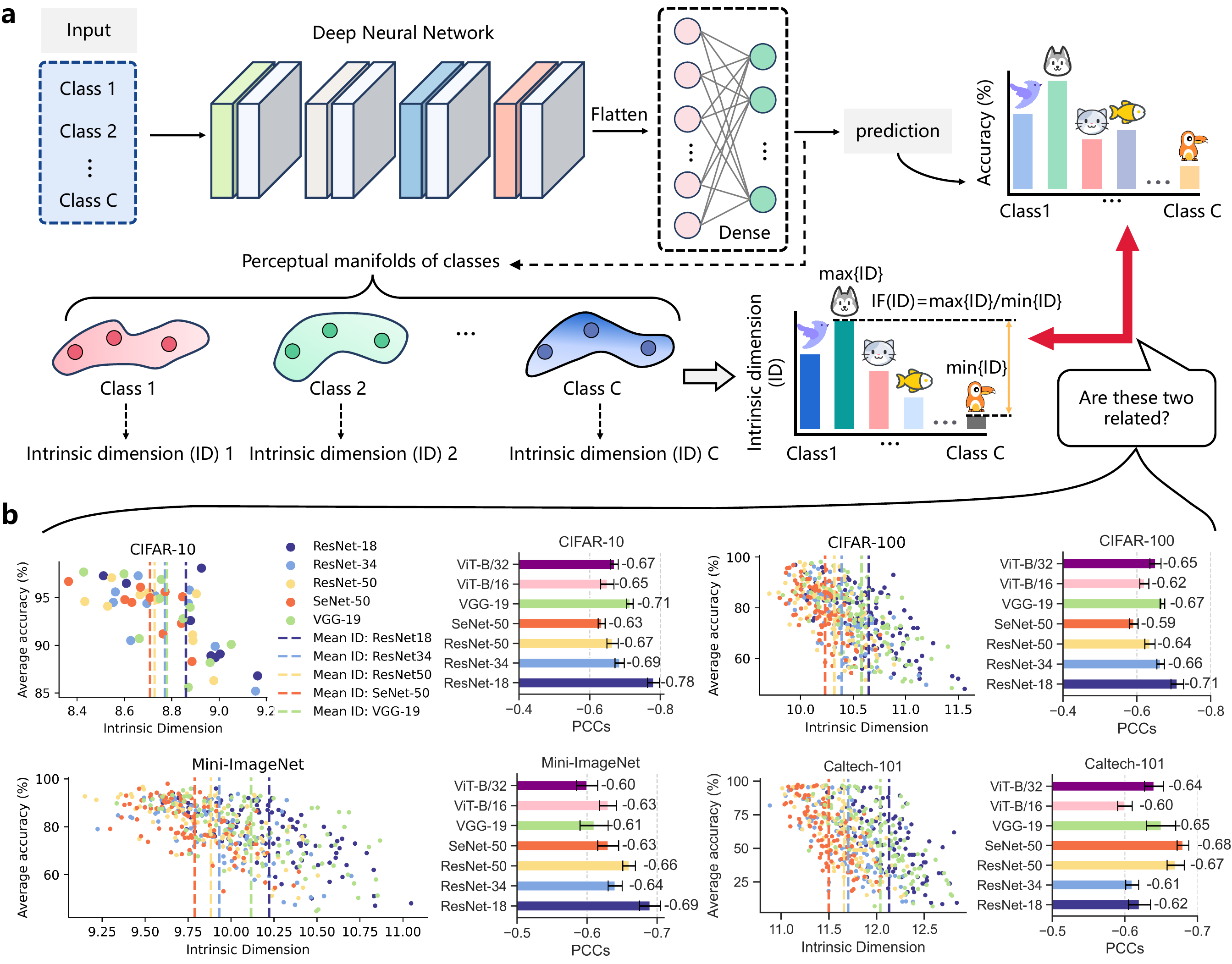}
\end{center}
\vspace{-0.4cm}
\caption{\textbf{The negative correlation between the intrinsic dimensions of class perceptual manifolds generated by the last hidden layer of the DNN and the class accuracy.} \textbf{a} Extracting image embeddings for each class from the last hidden layer of the DNN and calculating the intrinsic dimensions of class perceptual manifolds for each embedding set. \textbf{b} The scatter plot visualizes the distribution of class accuracy against the intrinsic dimensions of class perceptual manifolds, while the bar chart further illustrates their Pearson correlation coefficients (PCCs).}
\label{fig3}
\end{figure*}

\begin{figure*}[tb]
\begin{center}
\includegraphics[width=0.95\linewidth]{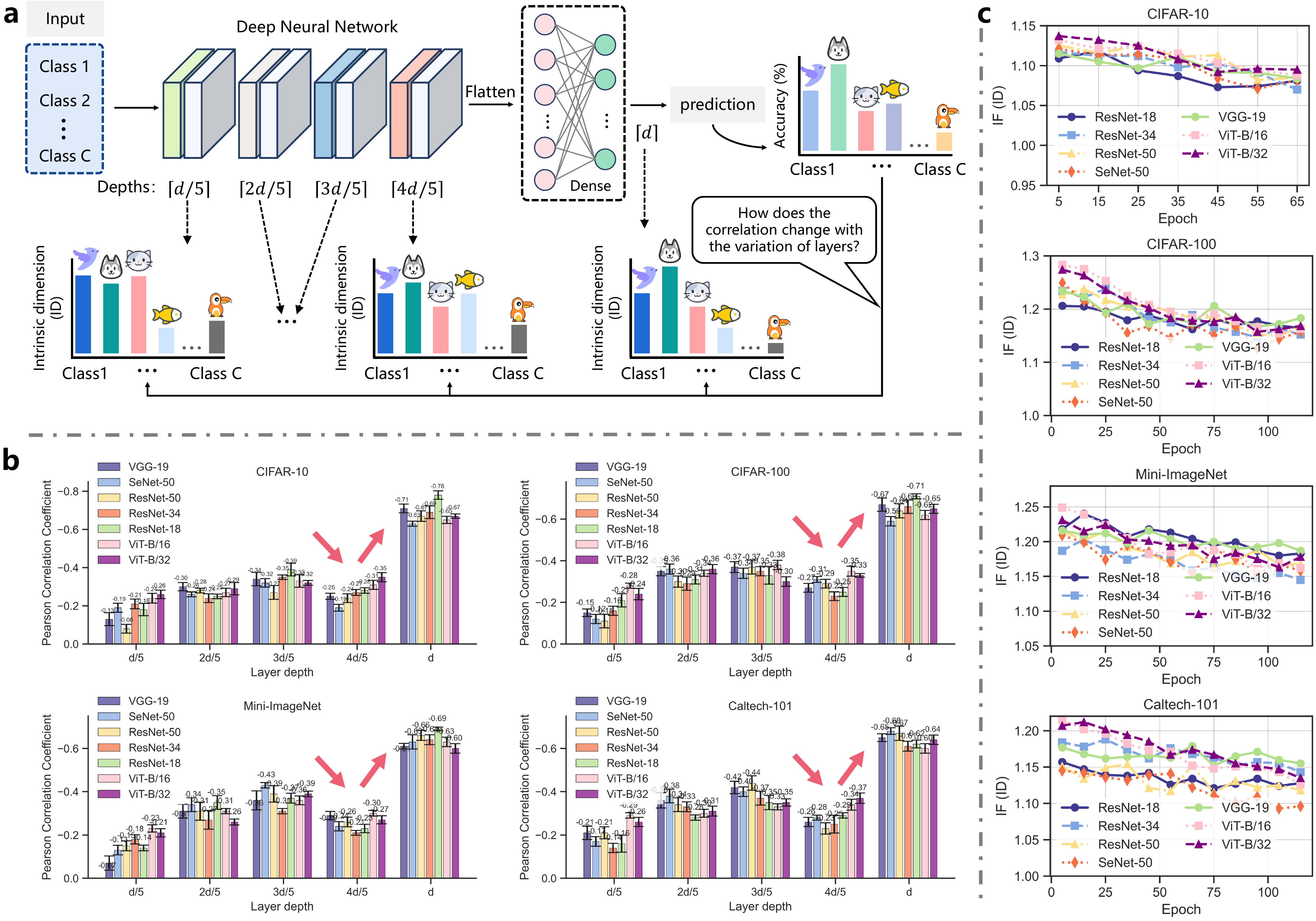}
\end{center}
\vspace{-0.15in}
\caption{\textbf{The impact of different layers of the DNN and the learning process on the correlation between the intrinsic dimensions of class perceptual manifolds and class accuracy.} \textbf{a} Extracting image embeddings for each class from different layers of well-trained DNNs and calculating the intrinsic dimensions of class perceptual manifolds. \textbf{b} The correlation between the intrinsic dimensions of class perceptual manifolds generated by different layers and class accuracy. \textbf{c} The existing optimization objectives are almost unable to reduce the imbalance of intrinsic dimensions among perceptual manifolds.}
\label{fig4}
\end{figure*}

\subsection{Learning Reduces the Intrinsic Dimension}

To use ID as a training objective or an additional constraint, it is necessary to explore how ID changes during the learning process, to enhance optimization objectives rather than conflicting with them. During the training of DNNs, we saved the model parameters every $10$ epochs and used these models to extract embeddings of the entire image dataset at the last hidden layer. Fig.\ref{fig2}c shows the IDs of perceptual manifolds at different training stages. We observed a clear trend: as training progresses, the IDs of perceptual manifolds gradually decreases. This result suggests that deep learning models tend to learn more concise data representations. Perceptual manifold with lower ID may imply that data points are more tightly clustered and concentrated in the embedding space, which helps the model learn more discriminative features. While this phenomenon aligns with our intuition, we have quantitatively revealed its existence for the first time through analysis, providing a solid empirical foundation for future research, rather than relying solely on qualitative analysis.

\subsection{ID Imbalance between Class Perceptual Manifolds Significantly Correlates with Model Fairness}

We have demonstrated that when considering the entire dataset as a single data manifold, there exists a negative correlation between the IDs of perceptual manifolds generated by the last hidden layer and the overall performance of the DNNs. Furthermore, we are curious whether the model's preferences for different categories are related to the IDs of the corresponding perceptual manifolds. We first define a measure for the imbalance of ID. As illustrated in Fig.\ref{fig3}a, for a $C$-classification task, let the IDs of the perceptual manifolds corresponding to the $C$ categories be $ID_1,ID_2,\dots,ID_C$. The imbalance of ID is defined as $IF(ID)=\max\{ID_1,ID_2,\dots,ID_C\}/\min\{ID_1,ID_2,\dots,ID_C\}$ \cite{ma5,cb_loss}. A higher $IF(ID)$ indicates a greater imbalance. Similarly, the model bias, $IF(ACC)$, is defined as the ratio of the highest class accuracy to the lowest class accuracy.

As shown in Fig.\ref{fig3}a, we extracted embeddings of each class of images generated by the last hidden layer of the DNNs and saved them separately. Subsequently, we estimated the IDs of the perceptual manifolds corresponding to each class embedding and calculated the Pearson correlation coefficients between them and the class accuracy. Fig.\ref{fig3}b illustrates the distribution of IDs of each perceptual manifold against its corresponding class accuracy. \textbf{We reached two key conclusions:} \textbf{(1)} For the same model, the IDs of perceptual manifolds corresponding to classes are negatively correlated with their accuracies. \textbf{(2)} For classes, as the complexity of the model structure increases, the IDs of the generated perceptual manifolds tend to decrease. These two conclusions provide a novel geometric perspective for explaining model fairness. Moreover, they offer a potential geometric constraint for mitigating model bias by balancing and reducing the IDs of perceptual manifolds corresponding to all classes.

Furthermore, we explored the correlation between the IDs of class perceptual manifolds generated by other layers of the DNN and class accuracy (as shown in Fig.\ref{fig4}a), demonstrated in Fig.\ref{fig4}b. Considering the classification model as a combination of representation network and classifier, \textbf{we found that:} \textbf{(1)} The correlation between the IDs of class perceptual manifolds in the last hidden layer and class accuracy is consistently higher than other layers. \textbf{(2)} There is no fixed pattern of correlation variation in the shallow layers, but at the end of the representation network, the correlation is always at a ``low point''. The first phenomenon determines the optimal application position for geometric constraints used to mitigate model bias, namely, the last hidden layer. The second phenomenon aligns with the motivation behind decoupled training paradigms \cite{ma1,Decoupled,BBN,ma2024geometric} in imbalance learning, where the model's preference is primarily caused by the classifier, while the representation network can obtain relatively unbiased image embeddings. These conclusions derived from a geometric perspective provide deeper insights into understanding the widespread issue of model unfairness.

\subsection{How Learning Affects ID Imbalance}

The bias of a trained DNN is closely related to the ID of the class perceptual manifold produced by the last hidden layer. This led us to wonder about how much the existing optimization objective affects the degree of imbalance in the IDs during the learning process. During the training of each classification model, we saved the model parameters every 10 epochs. Upon training completion, we extracted embeddings of each class image from the last hidden layer of all models. Then, we estimated the IDs of perceptual manifolds corresponding to each class, along with their imbalance levels. Fig.\ref{fig4}c illustrates the trend of the degree of imbalance of IDs with training epochs. We observed that all models attempt to reduce the ID imbalance during the learning process, but they lack strong constraints, resulting in weak effects. The high negative correlation between ID imbalance and model bias further substantiates this point.

\subsection{Balancing the ID of Class Perceptual Manifolds}

\begin{figure*}[!t]
\begin{center}
\includegraphics[width=\linewidth]{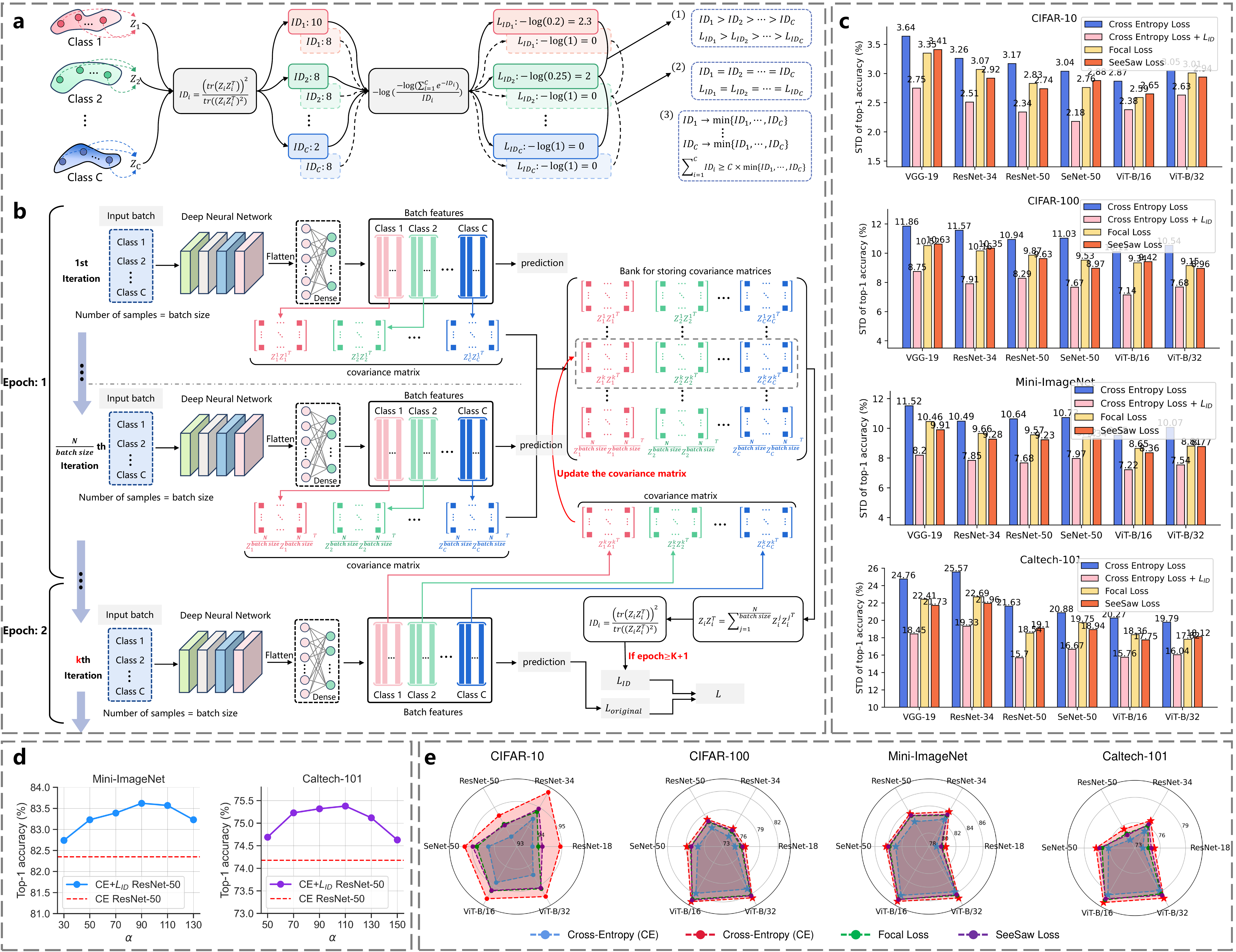}
\end{center}
\vspace{-0.1in}
\caption{\textbf{a\&b} The computation process of Intrinsic Dimension Regularization (IDR) and its corresponding training scheme. \textbf{a} We present two examples ($ID_1=10, ID_2=8, ID_C=2$ and $ID_1=ID_2=ID_C=8$) to verify whether $L_{ID}$ satisfies the first and second design principles. It can be observed that when $ID_1>ID_2>\dots>ID_C$, then $L_{ID_1}>L_{ID_2}>\dots>L_{ID_C}$. $L_{ID}$ is a decreasing function with respect to $ID$, thus fulfilling principle (1). When $ID_1=ID_2=\dots=ID_C$, then $L_{ID_1}=L_{ID_2}=\dots=L_{ID_C}=0$. Clearly, $L_{ID}$ conforms to principle (2). When $ID_i=\min\{ID_1,\dots,ID_C\}$, then $L_{ID_i}=0$, thus the purpose of $L_{ID}$ is to make the intrinsic dimensions of all class perceptual manifolds close to $min\{ID_1,\dots,ID_C\}$. Obviously, ${\textstyle \sum_{i=1}^{C}}ID_i \ge C\cdot \min\{ID_1,\dots,ID_C\}$, therefore $L_{ID}$ satisfies principle (3). \textbf{b} End-to-end training scheme. We construct a queue to store the covariance matrices generated at each iteration, which is capable of storing at most the covariance matrix generated in a complete epoch. As training progresses, the latest generated covariance matrix is continuously used to update the oldest covariance matrix in the queue. When applying $L_{ID}$, the stored covariance matrices in the queue can be used for its calculation. \textbf{c} The bias of multiple DNN models is significantly reduced by measuring the standard deviation of class accuracy, demonstrating the efficacy of intrinsic dimension regularization.\textbf{d} Using ResNet-50 as the backbone network, model performance under different values of $\alpha$ when employing $L_{ID}$. \textbf{e} IDR improves the overall performance of the model on multiple datasets.}
\label{fig5}
\end{figure*}

In this study, we propose \textbf{Intrinsic Dimension Regularization (IDR)}, which aims to improve the performance and fairness of DNNs by reducing the imbalance of IDs among class perceptual manifolds. In order to learn ID balanced and concise perceptual manifolds, \textbf{IDR needs to adhere to the following three principles:} 
\begin{itemize}
\item[(1)] The larger the ID of a class perceptual manifold, the stronger the penalty imposed on it. This principle is based on our experimental observation: reducing ID is often associated with an improvement in model performance.
\item[(2)] When IDs are balanced, the penalty intensity for each perceptual manifold is the same.
\item[(3)] The sum of IDs for all perceptual manifolds tends to decrease.
\end{itemize}

Given a $C$-classification task and a DNN, suppose the $C$-dimensional embedding of each class's images in the last hidden layer is represented as $Z_i=[z_i^1,\dots,z_i^{m_i}]\in\mathbb{R}^{C\times m_i}$, where $i=1,\dots,C$, and $m_i$ denotes the number of samples in class $i$. The intrinsic dimensionality of the perceptual manifold corresponding to class $i$ is estimated as $ID_i=\frac{(tr(Z_iZ_i^T))^2}{tr((Z_iZ_i^T)^2)}$ \cite{litwin2017optimal}. Intrinsic Dimension Regularization term is formally defined as $L_{ID}=  \sum_{i=1}^{C}-\log(\frac{ID_i^{-1}}{\max\{ID_1^{-1},\dots,ID_C^{-1}\}})$.

As shown in Fig.\ref{fig5}a, $L_{ID}$ conforms to three design principles. Considering the differentiability of the loss function, we use a smoothed form of the $\max$ function, resulting in the final form of $L_{ID}$: 
\begin{equation}
\begin{split}
L_{ID}^{smooth} &= \sum_{i=1}^{C}-\log\left(\frac{ID_i^{-1}}{\log\left( \sum_{i=1}^{C}e^{ID_i^{-1}}\right)}\right) 
\end{split}
\nonumber
\end{equation}

During model training, $L_{ID}$ can serve as a supplementary term to the existing loss function, typically formulated as $L=L_{original}+\frac{\log_{\alpha}epoch}{(\frac{L_{ID}^{smooth}}{L_{original}}).detach()}\times L_{ID}^{smooth}$, where $\alpha>1$. The purpose of the $(\frac{L_{ID}^{smooth}}{L_{original}}).detach()$-term is to keep the values of $L_{ID}$ and $L_{original}$ at the same scale. By adjusting $\log_{\alpha}epoch$, the trade-off between $L_{ID}$ and $L_{original}$ can be controlled, with the influence of $L_{ID}$ gradually increasing with the increase in epochs. The parameter $\alpha$ determines the rate of growth of $L_{ID}$. When $\alpha$ equals the epoch, $\log_{\alpha}epoch$ equals $1$. Therefore, a smaller $\alpha$ can cause the weight $\log_{\alpha}epoch$ to reach $1$ earlier. The regularization term form proposed in \cite{ma1} has proven effective, so we adopt a consistent formulation. However, the primary distinction lies in the optimization objectives: the objective in \cite{ma1} is to reduce and balance the curvature of perceptual manifolds, whereas the objective of this work is to reduce and balance their intrinsic dimensions. Additionally, to address the substantial memory requirements imposed by the training framework in \cite{ma1}, we propose approximating the global covariance matrix with multiple local covariance matrices, thereby eliminating the need to store image embeddings and significantly reducing storage costs.

As training progresses, the IDs of class perceptual manifolds continually shift, necessitating real-time updates. Re-extracting embeddings of the entire dataset at each iteration not only interrupts training but also greatly increases time costs.
Feature slow shift phenomenon \cite{Feature_slow_shift,ma3} indicates that as training progresses, the shift of sample embeddings becomes increasingly small, to the extent that historical versions of embeddings can approximate the current version. Inspired by this observation, we devise an end-to-end training scheme to apply IDR, as visualized in Fig.\ref{fig5}b. The key steps of this scheme are summarized as follows: 
\begin{itemize}
\item[(1)] In the first epoch of training, compute and store the covariance matrix of embeddings from each class at each iteration. Use $Z_i^j {Z_i^j}^T$ to denote the covariance matrix belonging to class $i$ at the $j$-th iteration. Thus, each class possesses $N/BS$ covariance matrices, where $N$ represents the total sample numbers of the dataset, and $BS$ denotes the batch size.

\item[(2)] From the second epoch to the $K$-th epoch, continuously update the oldest batch of covariance matrices using the covariance matrices newly computed at each iteration. The purpose of this step is to wait until the shift of embeddings becomes sufficiently small.

\item[(3)] Starting from the ($K+1$)-th epoch, we not only continue to update the covariance matrices but also utilize the latest $N/BS$ covariance matrices to approximate the overall covariance matrix $Z_i {Z_i}^T = \frac{1}{N} \sum_{j=1}^{N/BS} Z_i^j {Z_i^j}^T$. Through this process, we are able to estimate the intrinsic dimension of each class perceptual manifold and avoid interrupting the training. Subsequently, we compute and apply $L_{ID}$.
\end{itemize}

\begin{figure}[t]
\begin{center}
\includegraphics[width=\linewidth]{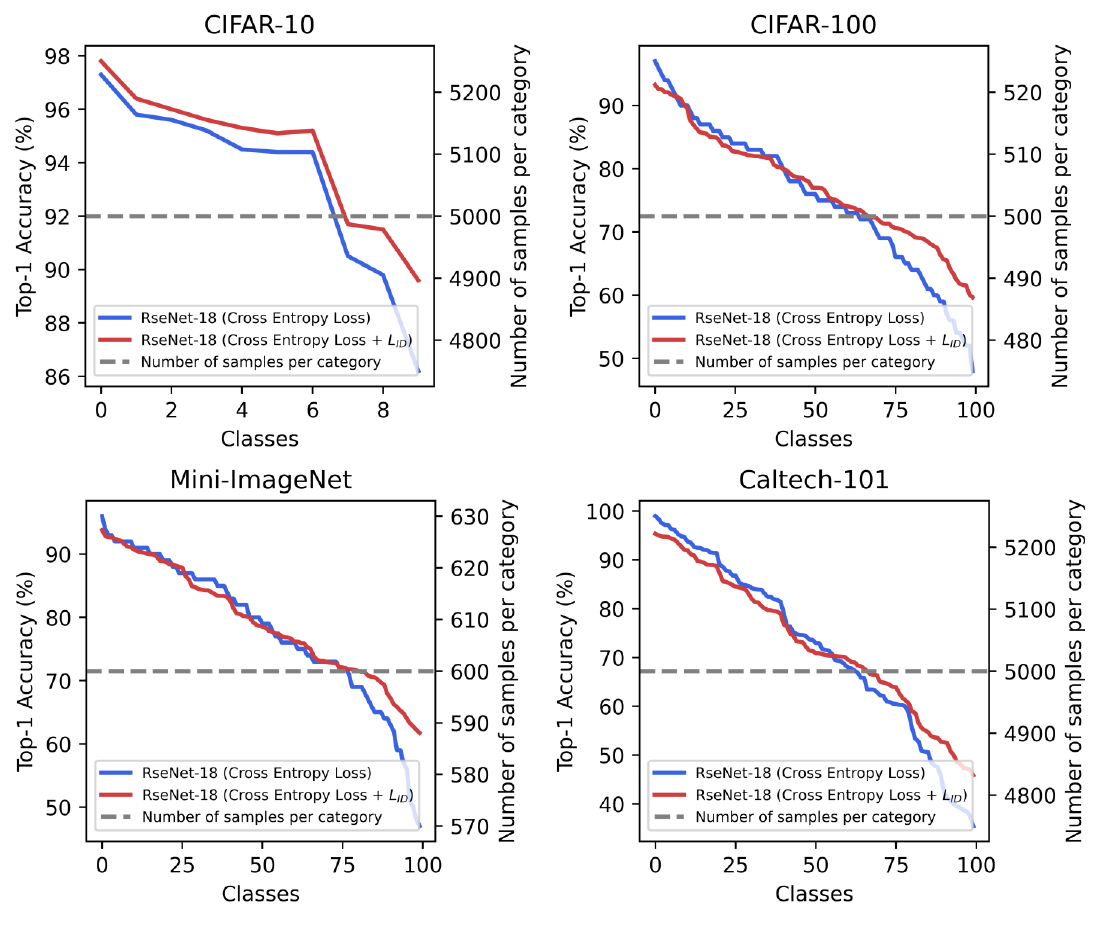}
\end{center}
\vspace{-0.2in}
\caption{Per-class accuracy across four datasets using ResNet-18 as the backbone network and cross-entropy as the classification loss, comparing results before and after applying IDR.}
\label{fig6}
\vskip -0.1in
\end{figure}

Fig.\ref{fig5}d illustrates the impact of different values of $\alpha$ on model performance. It can be observed that increasing $L_{ID}$ in the mid-to-late stages of training can more effectively enhance the overall performance of the model. We validated the effectiveness of IDR on four benchmark image datasets \cite{cifar,imagenet,caltech101} using DNN models based on both \textbf{convolutional and transformer architectures \cite{vgg,resnet,vit}}. Fig.\ref{fig5}c demonstrates the standard deviation of class accuracy on the test set before and after employing IDR. Fig.\ref{fig6} further illustrates the per-class accuracy with and without IDR, where \textbf{it is evident that incorporating \( L_{ID} \) significantly reduces the bias in deep neural networks.} Additionally, $L_{ID}$ comprehensively enhances the performance of multiple baseline models, particularly achieving good results on datasets with balanced sample sizes such as CIFAR-10, CIFAR-100, and Mini-ImageNet (see Fig.\ref{fig5}e). 

Compared to existing methods aimed at mitigating model bias, such as Focal Loss \cite{focal} and SeeSaw Loss \cite{seesaw}, our method consistently achieves superior overall performance across four class-balanced datasets. Notably, while Focal Loss and SeeSaw Loss only marginally reduce model bias, IDR demonstrates a clear advantage. Importantly, Focal Loss and SeeSaw Loss differ fundamentally from our approach, as they adjust sample weights based on model predictions, while our approach derives class difficulty directly from the data, providing a new perspective and optimization target for bias mitigation. Unlike Focal Loss and SeeSaw Loss, which offer no new optimization targets, IDR introduces a novel objective by focusing on class-level intrinsic dimensions.
This performance of IDR is exciting because in the past, researchers have not paid much attention to model fairness on datasets with balanced sample sizes.
In conclusion, this work provides a novel geometric perspective and tool for studying the fairness of deep neural networks, prompting AI researchers to pay attention to a broader range of imbalance phenomena.

\section{Conclusion and Future Works}
In this study, we first explored the association between the intrinsic dimension of perceptual manifold and the overall performance and fairness of the model from the geometric perspective of data classification. We then proposed intrinsic dimension regularization (IDR) to reduce and balance the intrinsic dimensions of perceptual manifolds corresponding to each class, thereby mitigating model bias. Most previous studies have assumed that classes with fewer samples are weak classes. Following this assumption, DNN models trained on datasets with perfect class balance should not exhibit bias, or in other words, DNNs should be entirely fair. However, recent research indicates that classes with fewer samples are not always challenging to learn, and models trained on class-balanced datasets may even exhibit significant bias. A significant departure from them is that measuring model fairness from a geometric perspective is no longer exclusive to long-tailed datasets but may also be applicable to datasets with perfectly balanced samples. 
Our study has opened up a geometric perspective on researching the fairness of deep neural networks. From this perspective, all geometric characteristics concerning the complexity of perceptual manifolds might be predictive of model fairness. Therefore, extending a range of geometric measures for evaluating model fairness is a direction for our future research.

{
\small
\bibliographystyle{IEEEtran}
\bibliography{egbib}
}

\ifCLASSOPTIONcaptionsoff
  \newpage
\fi

\end{document}